\documentclass[letterpaper, 10 pt, conference]{ieeeconf}  

\IEEEoverridecommandlockouts                              

\usepackage{algorithm}
\usepackage{algorithmic}
\usepackage{amsmath}
\usepackage{multirow}
\usepackage{float}
\usepackage{graphicx}
\usepackage{booktabs}
\usepackage{amssymb}

\title{\LARGE \bf
QueryGaussian: Scalable and Training-Free Open-Vocabulary 3D Instance Retrieval
}

\author{Xiuyuan Zhu$^{1,2}$, Ke Lu$^{1,3}$, Zijie Yang$^{2}$, Chao Yue$^{1,2}$, Jian Xue$^{1}$, Dongming Zhang$^{2}$%
\thanks{This work was supported by the National Natural Science Foundation of China (62521007, 62032022, 62320106007). (\emph{Corresponding author: Jian Xue})}%
\thanks{$^{1}$University of Chinese Academy of Sciences, Beijing, China}%
\thanks{$^{2}$State Key Laboratory of Communication Content Cognition, Beijing, China}%
\thanks{$^{3}$Peng Cheng Laboratory, Shenzhen, China}%
}

\begin{document}

\maketitle
\thispagestyle{empty}
\pagestyle{empty}

\begin{abstract}
Efficiently retrieving specific 3D instances from large-scale scenes via natural language prompts remains a formidable challenge in multimedia analysis. Existing approaches predominantly adhere to a ``scene-level embedding'' paradigm, which necessitates distilling high-dimensional semantic features into every 3D primitive. This strategy suffers from a fundamental architectural bottleneck: the memory and computational costs scale linearly with scene complexity, inevitably triggering Out-Of-Memory (OOM) failures in city-scale environments. To dismantle this barrier, we propose QueryGaussian, a training-free framework enabling expeditious and scalable open-vocabulary 3D instance retrieval. Diverging from holistic semantic distillation, QueryGaussian employs a novel Instance-Level Query Mechanism that rigorously decouples semantic understanding from geometric representation. Specifically, we leverage pre-trained 2D vision models to interpret user prompts and lift segmentation masks into 3D via a concurrent maximum-weight association strategy, ensuring strict semantic-visual consistency. To mitigate projection ambiguity, we introduce a temporal fusion module with multi-stage adaptive density clustering. Experimental results demonstrate that QueryGaussian not only matches the accuracy of state-of-the-art methods but also delivers a decisive efficiency leap—reducing GPU memory usage by over 70\% and accelerating inference by 180$\times$. Crucially, QueryGaussian enables expeditious instance retrieval on city-scale scenes (containing tens of millions of Gaussians) on consumer-grade hardware.
\end{abstract}


\section{Introduction}

3D Gaussian Splatting (3DGS) \cite{kerbl3Dgaussians} has recently emerged as an effective representation for multimedia content \cite{kerbl3Dgaussians, Fang2024MiniSplattingRS, franke2024trips, song2024sa, yan2024multi}. By representing a scene with explicit 3D Gaussian primitives and projecting them onto 2D image planes, 3DGS combines geometric structure with high-fidelity rendering. This formulation has supported a wide range of applications, including virtual and augmented reality \cite{10896112}, autonomous driving \cite{zhou2024drivinggaussian, chen2025omnire}, and large-scale urban reconstruction \cite{Wu_2024_CVPR, Yang_2024_CVPR, Huang_2024_CVPR}. As 3D scenes become larger and more complex, an important problem is how to retrieve and localize specific 3D instances from arbitrary-scale scenes using natural language queries.

This problem is often studied as open-vocabulary 3D instance retrieval \cite{wu2024opengaussian, Chacko_2025_WACV, Shi_2024_CVPR, Qin_2024_CVPR}. It remains challenging because of occlusion, scale variation, and the large volume of 3D data. Traditional 3D representations such as point clouds (e.g., PointGroup \cite{jiang2020pointgroup}) and voxels (e.g., VoxelEmbed \cite{10.1007/978-3-030-87589-3_45}) are often computationally manageable, but they tend to lose fine-grained detail. In contrast, implicit neural representations such as LERF \cite{Kerr_2023_ICCV}, 3D-OVS \cite{liu2023weakly}, and TensoRF \cite{Chen2022ECCV} can model scenes with high fidelity, yet their retrieval efficiency is limited. These limitations make scalable and efficient 3D instance retrieval still an open problem.

Recent methods that extend 3DGS to open-vocabulary scene understanding mostly follow a \textit{scene-level embedding} paradigm \cite{wu2024opengaussian, Chacko_2025_WACV, Shi_2024_CVPR, Qin_2024_CVPR}. In this pipeline, semantic features are extracted from dense multi-view images and then distilled into 3D Gaussians through additional optimization or retraining. Retrieval can only be performed after this offline process. Although effective on small scenes, this design has several drawbacks. First, extracting and fusing high-dimensional semantic features over the entire scene introduces substantial preprocessing cost. Second, because semantic information is baked into the scene representation, adapting to a new vocabulary or a new scene often requires rebuilding the semantic representation. Third, and most importantly, memory usage grows with scene scale, which often causes Out-Of-Memory (OOM) failures on consumer GPUs for city-scale scenes. These issues limit the practicality of scene-level embedding methods in large-scale settings.

This paper presents \textbf{QueryGaussian}, a training-free framework for efficient open-vocabulary 3D instance retrieval. Instead of processing the full scene in advance, QueryGaussian follows an on-demand retrieval strategy. Given a text query, it uses pre-trained 2D vision models~\cite{liu2023grounding,ravi2024sam2segmentimages,kirillov2023segany,ren2024grounded} to identify target regions in rendered views, and then lifts these 2D results into 3D through a lightweight maximum-weight association strategy. This design separates semantic understanding from scene representation and therefore avoids scene-specific semantic distillation or retraining. As a result, QueryGaussian can perform zero-shot retrieval directly on pre-trained 3DGS scenes of arbitrary scale.

To improve retrieval quality, QueryGaussian further includes an \textbf{Adaptive Instance Clustering with Temporal Fusion} module. The lifted 3D candidates may contain noise caused by projection ambiguity, occlusion, and reconstruction artifacts. To address this issue, temporal consistency across views is used for incremental filtering, followed by multi-stage density clustering to refine the final instance set. In this way, the method remains lightweight while improving spatial precision.

Overall, QueryGaussian provides a simple and scalable alternative to scene-level semantic embedding. By reusing native 3DGS parameters and rasterization byproducts, it avoids the cost of dense semantic feature distillation and significantly reduces memory overhead. Experiments show that QueryGaussian achieves competitive retrieval accuracy while scaling to city-scale scenes with tens of millions of Gaussians on a single GPU, where prior scene-level embedding methods fail due to resource limitations.

The main contributions are summarized as follows:
\begin{itemize}
    \item A training-free framework, QueryGaussian, for open-vocabulary 3D instance retrieval on 3DGS scenes, designed to support efficient querying from small indoor scenes to city-scale environments.

    \item An instance-level query mechanism that avoids scene-wide semantic feature distillation and performs direct 2D-to-3D association from pre-trained 2D segmentation results.

    \item An adaptive refinement module that combines temporal fusion and multi-stage density clustering to suppress outliers and improve instance localization quality.

    \item Extensive experiments showing that QueryGaussian achieves competitive retrieval accuracy while reducing GPU memory usage by more than $70\%$ and accelerating inference by up to $180\times$ compared with prior methods.
\end{itemize}

\begin{figure*}[!ht]
  \includegraphics[width=\textwidth]{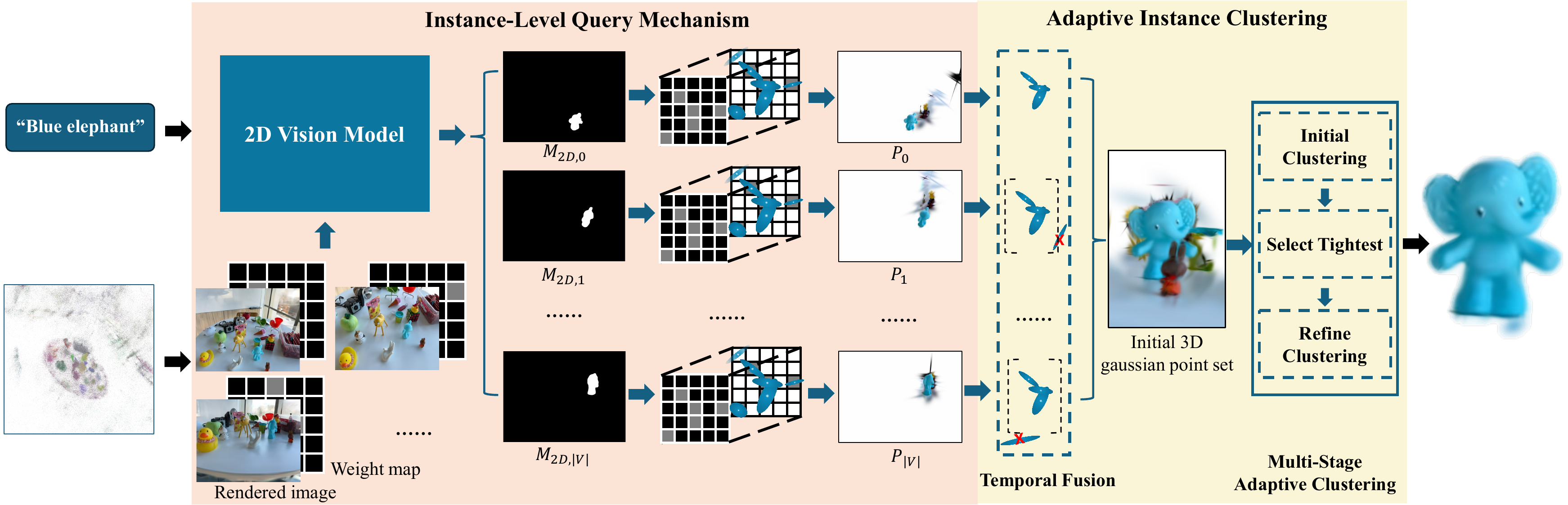}
  \caption{
    Overview of QueryGaussian. Given a 3DGS scene and a text query, the framework first renders multi-view images and records per-pixel maximum-weight maps. The instance-level query module then applies an open-vocabulary 2D segmentation model to obtain query-related masks and lifts them into an initial 3D point set through maximum-weight association. Finally, adaptive instance clustering with temporal fusion removes outliers across views and refines the result into a clean 3D instance segmentation.
    }
  \label{fig:framework}
\end{figure*}

\section{Related Work}

\subsection{Efficient 3D Scene Representation}
Neural Radiance Fields (NeRF) \cite{10.1145/3503250} pioneered high-quality implicit 3D scene representation but suffered from slow inference and high computational costs. While subsequent iterations optimized sampling and training speeds \cite{mueller2022instant, Garbin_2021_ICCV, yu2022plenoxels, LIU2025102752}, the introduction of 3D Gaussian Splatting (3DGS) \cite{kerbl3Dgaussians} marked a paradigm shift. By utilizing explicit point primitives and a differentiable rasterization pipeline, 3DGS achieves real-time rendering speeds and superior training efficiency.

Crucially for this work, the explicit nature of 3DGS has facilitated its expansion to large-scale environments. Recent methods like VastGaussian \cite{lin2024vastgaussian} and CityGaussian \cite{liu2025citygaussian} have successfully scaled 3DGS to city-level scenes via parallel training and Level-of-Detail (LoD) techniques \cite{huang2025hierarchicalcompressiontechnique3d, ren2024octree}. Additionally, dynamic scene modeling has been advanced by DrivingGaussian \cite{zhou2024drivinggaussian} and StreetGaussians \cite{yan2024street}. These developments provide the foundation for large-scale 3D content creation, yet efficient \textit{retrieval} within such massive scenes remains an unaddressed bottleneck.

\subsection{Open-Vocabulary 3D Semantics and Retrieval}
Recent works have attempted to bridge 3DGS with open-vocabulary semantics, primarily following a \textit{feature distillation} paradigm. Methods such as LangSplat \cite{Qin_2024_CVPR} and LEGaussians \cite{Shi_2024_CVPR} leverage pre-trained Vision Models (e.g., CLIP \cite{2021Learning}, SAM \cite{kirillov2023segany,ravi2024sam2}) to extract dense semantic maps from multi-view images. These features are then distilled into the 3D Gaussians via auto-encoders or quantization, constructing a queryable semantic field. OpenGaussian \cite{wu2024opengaussian} further extends this to point-level understanding by associating 3D points with 2D CLIP features via a discretized codebook. Similarly, Gaussian Grouping \cite{gaussian_grouping} introduces identity encoding to enforce 3D spatial consistency, enabling simultaneous reconstruction and segmentation.

Despite their success on object-centric or room-scale datasets, these "Scene-Level Embedding" approaches share critical limitations that hinder their application in multimedia retrieval. First, they require **coupled reconstruction or retraining**, meaning they cannot be directly applied to pre-existing 3D assets without expensive optimization. Second, the **computational and memory overhead** for distilling high-dimensional features across millions of points is prohibitive, often leading to Out-Of-Memory (OOM) failures in large-scale scenes \cite{liu2025citygaussian}.

In contrast, our proposed QueryGaussian shifts from global semantic embedding to **instance-level retrieval**. By strictly separating scene representation from semantic understanding, our method performs zero-shot queries on arbitrary-scale 3DGS scenes without any training or parameter tuning. This design ensures extremely high computational efficiency and scalability, making it uniquely suitable for city-scale retrieval tasks.

\section{Method}

The overall framework of QueryGaussian is shown in Figure~\ref{fig:framework}. Given a pre-trained 3DGS scene and a natural language query, the method performs open-vocabulary 3D instance retrieval in a training-free manner. The pipeline consists of three stages:
(1) \textbf{Render-and-Map}: the system renders multi-view images and simultaneously records a per-pixel maximum-weight map, which stores the index of the Gaussian with the largest contribution to each pixel;
(2) \textbf{Instance-Level Query Mechanism}: an open-vocabulary 2D segmentation model predicts query-relevant masks, which are then lifted to 3D through the maximum-weight map to form an initial candidate set;
(3) \textbf{Adaptive Instance Clustering with Temporal Fusion}: a refinement module combines cross-view consistency filtering and multi-stage density clustering to remove noise and outliers, producing a compact 3D instance representation.

\subsection{Preliminary: 3D Gaussian Splatting}
\label{sec:preliminary_3dgs}

3D Gaussian Splatting (3DGS) \cite{kerbl3Dgaussians} represents a scene as a set of explicit 3D Gaussian primitives. Each primitive is parameterized by its center position $\mathbf{p} \in \mathbb{R}^3$, covariance determined by a rotation quaternion $\mathbf{q} \in \mathbb{R}^4$ and a scaling factor $\mathbf{s} \in \mathbb{R}^3$, opacity $\alpha \in [0,1]$, and view-dependent color $\mathbf{c}$, typically represented with spherical harmonics.

Given a target view, 3DGS renders an image through differentiable rasterization. For a pixel $p$, its rendered color $C(p)$ is computed by accumulating the contributions of the $N$ ordered Gaussians intersecting the pixel:
\begin{equation}
    C(p) = \sum_{i=1}^{N} \mathbf{c}_i \underbrace{\alpha_i \prod_{j=1}^{i-1} (1-\alpha_j)}_{\omega_i},
    \label{eq:alpha_blending}
\end{equation}
where $\omega_i$ denotes the blending weight of the $i$-th Gaussian at pixel $p$. This explicit weighting formulation is central to our method, because it identifies which Gaussian contributes most strongly to the visible appearance at each pixel. We use this property to establish a direct 2D-to-3D association without introducing auxiliary semantic embeddings.

\subsection{Instance-Level Query Mechanism}
\label{sec:elevating_2d_to_3d}

Existing scene-level embedding methods \cite{wu2024opengaussian, Chacko_2025_WACV} distill semantic features for the entire scene before retrieval. This design is expensive and scales poorly with scene size. In contrast, our instance-level query mechanism performs retrieval on demand. It decouples semantic understanding from scene representation and enables direct querying on pre-trained 3DGS scenes without additional training.

\subsubsection{Open-Vocabulary 2D Semantic Acquisition}

We use Grounded-SAM \cite{liu2023grounding, ren2024grounded} as the 2D semantic parser. Given a text prompt $T$ and a rendered image $I_v$ from view $v$, the model predicts a binary mask
$
M_{\text{2D},v} \in \{0,1\}^{H \times W}.
$
This step relies entirely on the open-vocabulary capability of large-scale 2D foundation models and does not require any 3D-specific supervision.

\subsubsection{Max-Weight 2D-to-3D Association}

To lift the 2D mask $M_{\text{2D},v}$ into 3D, we must resolve the ambiguity that each pixel may correspond to multiple Gaussians along the viewing ray. A depth-based assignment is not reliable in this setting, because low-opacity artifacts or floaters may lie in front of the actual visible surface. Instead, we assign the pixel to the Gaussian with the largest rendering contribution.

Formally, for each pixel $p$, we define the dominant Gaussian index as
\begin{equation}
    \operatorname{idx}_{\max}(p) =
    \operatorname*{arg\,max}_{i \in \mathcal{N}(p)} \omega_{i,p},
    \quad
    \text{where }
    \omega_{i,p} = \alpha_i \prod_{j=1}^{i-1}(1-\alpha_j),
    \label{eq:max_weight_gaussian}
\end{equation}
and $\mathcal{N}(p)$ is the ordered set of Gaussians intersecting the ray of pixel $p$. We refer to the resulting index buffer as the \textbf{maximum-weight map}. It is generated together with the rendering pass and introduces negligible additional overhead.

For each active mask pixel satisfying $M_{\text{2D},v}[p]=1$, we retrieve the corresponding Gaussian $G_{\operatorname{idx}_{\max}(p)}$ from the maximum-weight map and add it to the candidate set $P_{\text{inst}}$. This operation directly lifts the 2D segmentation result to 3D while suppressing many irrelevant semi-transparent foreground artifacts.

\subsubsection{Complexity Analysis}

Scene-level embedding methods typically require querying or storing a $D$-dimensional feature for all $N_{\text{scene}}$ Gaussians, leading to semantic processing cost on the order of $O(N_{\text{scene}} \times D)$. In our method, the rasterization stage still depends on scene geometry, but the semantic association step itself operates purely in image space. Its cost is therefore bounded by $O(N_{\text{pixels}})$ for each rendered view. This design avoids dense semantic processing over all Gaussians and is substantially more favorable for large scenes.

\renewcommand{\algorithmicrequire}{\textbf{Input:}}
\renewcommand{\algorithmicensure}{\textbf{Output:}}
\begin{algorithm}
\caption{Instance-Level Query Mechanism}
\label{alg:elevating}
\begin{algorithmic}[1]
\REQUIRE Text prompt $T$, set of viewpoints $V$
\ENSURE Candidate instance set $P_{\text{inst}}$
\STATE $P_{\text{inst}} \leftarrow \emptyset$
\FOR{$v \in V$}
    \STATE $I_v, \text{MaxWeightMap}_v \leftarrow \operatorname{RenderWithMaxWeight}(v)$
    \STATE $M_{\text{2D},v} \leftarrow \operatorname{GroundedSAM}(I_v, T)$
    \FOR{$p \in \operatorname{Pixels}(I_v)$}
        \IF{$M_{\text{2D},v}[p] = 1$}
            \STATE $\text{idx} \leftarrow \text{MaxWeightMap}_v[p]$
            \STATE $P_{\text{inst}} \leftarrow P_{\text{inst}} \cup \{G_{\text{idx}}\}$
        \ENDIF
    \ENDFOR
\ENDFOR
\RETURN $P_{\text{inst}}$
\end{algorithmic}
\end{algorithm}

\subsection{Adaptive Instance Clustering with Temporal Fusion}
\label{sec:adaptive_clustering}

The initial candidate set $P_{\text{inst}}$ obtained from 2D-to-3D lifting inevitably contains noise, including floaters and reconstruction artifacts. To refine this set, we introduce a module that combines incremental cross-view filtering with multi-stage density clustering.

\subsubsection{Incremental Temporal Filtering}

We use multi-view consistency to suppress transient outliers. Let $P_t$ denote the point set obtained from the $t$-th view, and let $F_{t-1}$ be the accumulated set from previous views, with $F_0 = P_0$. For each candidate point $p \in P_t$, we measure its consistency with the historical set $F_{t-1}$ using the average Euclidean distance:
\begin{equation}
    d_{\text{avg}}(p, F_{t-1}) =
    \frac{1}{|F_{t-1}|}
    \sum_{q \in F_{t-1}} \|\mathbf{p}_p - \mathbf{p}_q\|_2.
    \label{eq:avg_dist_history}
\end{equation}
Points with large deviations are likely to be outliers. To adapt to scale variation across views, we define a dynamic threshold
\begin{equation}
    \tau_t = \eta \cdot \mathop{\operatorname{median}}_{p' \in P_t}
    \left\{ d_{\text{avg}}(p', F_{t-1}) \right\},
    \label{eq:dynamic_threshold}
\end{equation}
where $\eta$ is a sensitivity factor. The filtered set is then
\begin{equation}
    P_t^{\text{filtered}} =
    \mathcal{F}_{\text{cap}}
    \left(
    \left\{ p \in P_t \mid d_{\text{avg}}(p, F_{t-1}) \le \tau_t \right\}
    \right),
    \label{eq:filtered_points}
\end{equation}
where $\mathcal{F}_{\text{cap}}(\cdot)$ applies a rejection-rate cap to avoid over-pruning. The global set is updated as
$
F_t = F_{t-1} \cup P_t^{\text{filtered}},
$
and the final fused set is denoted by $F_{\text{final}}$.

\subsubsection{Multi-Stage Adaptive Density Clustering}

Although temporal fusion removes many unstable points, the fused set $F_{\text{final}}$ may still contain structural noise. We therefore apply a multi-stage clustering procedure based on DBSCAN \cite{10.5555/3001460.3001507}. DBSCAN is well suited for this problem because it handles irregular cluster shapes and varying densities without requiring the number of clusters in advance.

\textbf{Phase 1: Initial coarse clustering.}
We first apply DBSCAN with global parameters $\theta_{\text{initial}}$ to $F_{\text{final}}$ and extract the largest connected cluster as an initial proposal:
\begin{equation}
    C_1 = \operatorname{LargestCluster}\bigl(\operatorname{DBSCAN}(F_{\text{final}}, \theta_{\text{initial}})\bigr).
\end{equation}

\textbf{Phase 2: Tightness-driven selection.}
The initial cluster may still include loosely connected structures. We therefore refine $C_1$ by selecting its most geometrically compact subset. Specifically, we define a tightness criterion based on average pairwise distance and use $\operatorname{SelectTightest}$ to obtain the sub-cluster $C_{\text{tight}} \subseteq C_1$ with the strongest internal coherence.

\begin{figure*}[!htp]
  \centering
  \includegraphics[width=\textwidth]{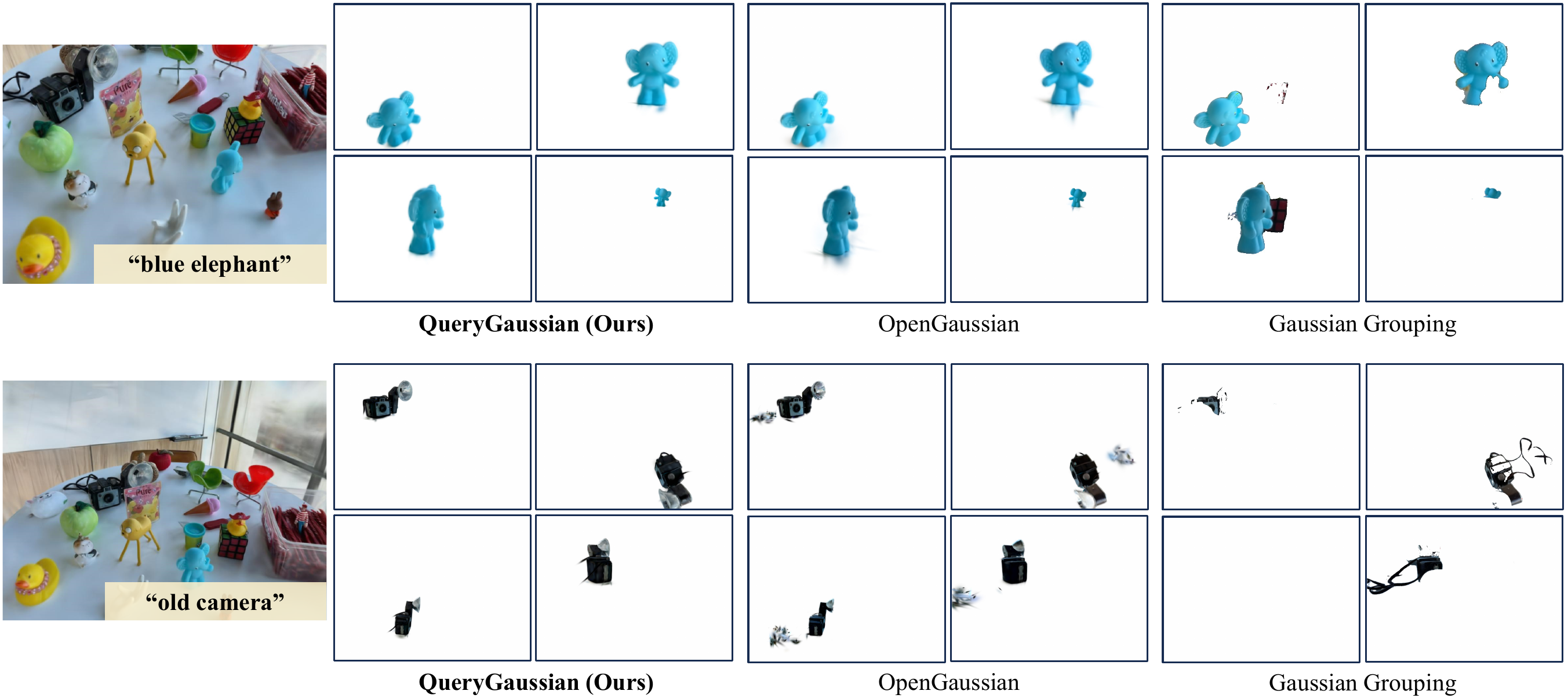}
  \caption{Qualitative comparisons on the small-scale indoor scene dataset. QueryGaussian produces cleaner segmentation with fewer floaters and noise artifacts than the baseline methods.}
  \label{fig:lerf}
\end{figure*}

\textbf{Phase 3: Fine-grained refinement.}
Finally, we apply a second DBSCAN pass to $C_{\text{tight}}$ using stricter and locally adapted parameters $\theta_{\text{refine}}$:
\begin{equation}
    C_{\text{final}} =
    \operatorname{LargestCluster}\bigl(\operatorname{DBSCAN}(C_{\text{tight}}, \theta_{\text{refine}})\bigr).
    \label{eq:refinement_cluster}
\end{equation}
This coarse-to-fine design progressively removes residual boundary noise and produces a cleaner instance segmentation, as illustrated in Figure~\ref{fig:lerf}.

\section{Experiments}

\subsection{Experimental Settings}

\noindent\textbf{Datasets and benchmark protocol.}
We evaluate both fine-grained retrieval accuracy and large-scale scalability on a curated benchmark with five scenes from two regimes. The benchmark includes three small-scale indoor scenes from LERF \cite{lerf2023}, each containing roughly $10^5$ Gaussians, and two large-scale outdoor scenes from MatrixCity \cite{li2023matrixcity} and Rubble \cite{Turki_2022_CVPR}, each containing roughly $10^7$ Gaussians. Because the original annotations are not tailored to single-instance retrieval, we re-annotate all scenes and keep only clearly visible, unambiguous instances that match specific text queries. This protocol makes IoU a more reliable measure of complete instance retrieval rather than ambiguous region matching.

\noindent\textbf{Implementation details.}
We use Grounded-SAM \cite{liu2023grounding, ren2024grounded} as the open-vocabulary 2D segmentation backend. QueryGaussian is implemented in PyTorch and evaluated on a single NVIDIA RTX 4090 GPU with 24GB VRAM. We directly use official pre-trained 3DGS checkpoints without any additional scene-specific optimization. Indoor scenes are reconstructed with 30k iterations ($\sim$100k Gaussians), while outdoor scenes contain more than 10 million Gaussians. Unless otherwise specified, QueryGaussian requires about 20 seconds per query and peaks at about 7GB GPU memory.

\subsection{Ablation Studies}
\label{sec:ablation}

We ablate three aspects of the method: the clustering design, robustness to noisy 2D masks, and dependence on 3DGS reconstruction quality.

\subsubsection{Effect of Multi-Stage Adaptive Clustering}
\label{sec:ablation_clustering}

We replace our multi-stage clustering module with a single-stage HDBSCAN baseline \cite{McInnes2017}, while keeping all other components unchanged.

Table~\ref{tab:ablation_clustering} shows that single-stage clustering remains usable on small scenes but breaks down on large scenes, where mIoU drops to 0.0392. In contrast, the proposed multi-stage design performs well in both regimes. This result shows that the clustering module is important not only for final accuracy but also for large-scale robustness.

\begin{table}[htb]
\centering
\caption{Ablation on clustering strategy. Multi-stage clustering is important for large-scale scenes.}
\label{tab:ablation_clustering}
\resizebox{0.9\linewidth}{!}{
\begin{tabular}{lcc}
\toprule
Clustering Strategy & Small-scale Scene & Large-scale Scene \\
\midrule
Single-stage (HDBSCAN) & 0.4164 & 0.0392 \\
\textbf{Multi-stage Adaptive (Ours)} & \textbf{0.6317} & \textbf{0.7676} \\
\bottomrule
\end{tabular}
}
\end{table}

\subsubsection{Robustness to Segmentation Quality}
\label{sec:ablation_robustness_clustering}

We test robustness to imperfect 2D masks in two settings: replacing the default SAM 2.1 Large with SAM 2.1 Tiny to simulate over-segmentation, and using artificially degraded masks to simulate under-segmentation.

Table~\ref{tab:ablation_clustering_robustness} shows that when the mask is noisy, the lifted point count increases to 105.6\% of the baseline, but the final mIoU remains unchanged after refinement. When the mask is incomplete, only 85.0\% of the baseline points are retained, yet the final mIoU still preserves 97.2\% of the baseline. These results indicate that the refinement stage can effectively remove added noise and preserve the core structure under missing observations.

\begin{table}[ht]
\centering
\caption{Robustness to different mask qualities. ``Initial Points'' measures the amount of lifted points relative to the baseline, and ``Final mIoU'' is reported after adaptive clustering.}
\label{tab:ablation_clustering_robustness}
\resizebox{0.95\linewidth}{!}{
\begin{tabular}{lcc}
\toprule
Segmentation Source & Initial Point Count (Rel. \%) & Final mIoU Retention (Rel. \%) \\
\midrule
SAM 2.1 Large (Baseline) & 100.0 & 100.0 \\
SAM 2.1 Tiny & 105.6 & 100.0 \\
Artificially Degraded Mask & 85.0 & 97.2 \\
\bottomrule
\end{tabular}
}
\end{table}

\subsubsection{Impact of 3DGS Reconstruction Quality}
\label{sec:ablation_3dgs_scene}

Since QueryGaussian does not retrain the 3D scene, its upper bound depends on the quality of the underlying 3DGS reconstruction. We evaluate the same scene reconstructed with 20k and 30k training iterations.

As shown in Table~\ref{tab:ablation_3dgs_scene}, the higher-quality reconstruction yields a clear gain, improving mIoU from 0.4810 to 0.6317. This result confirms that better scene geometry reduces ambiguity in 2D-to-3D lifting and provides a stronger basis for retrieval.

\begin{table}[htb]
    \centering
    \caption{Impact of 3DGS reconstruction quality on retrieval performance.}
    \label{tab:ablation_3dgs_scene}
    \resizebox{0.8\linewidth}{!}{
    \begin{tabular}{lc}
        \toprule
        3DGS Scene Quality & mIoU \\
        \midrule
        Low-Quality (20k iterations) & 0.4810 \\
        \textbf{High-Quality (30k iterations)} & \textbf{0.6317} \\
        \bottomrule
    \end{tabular}
    }
\end{table}

\begin{table*}[!ht]
\caption{IoU comparison on small-scale indoor and large-scale outdoor scenes. ``-'' indicates that the method failed on the large-scale benchmark due to GPU out-of-memory.}
\label{tab:iou_results}
\resizebox{\textwidth}{!}{
\begin{tabular}{l|cccc|ccc}
\toprule
\multirow{2}{*}{Methods} & \multicolumn{4}{c|}{Small-scale Indoor Scene} & \multicolumn{3}{c}{Large-scale Outdoor Scene} \\
\cmidrule(lr){2-5} \cmidrule(lr){6-8}
& Figurines & Ramen & Teatime & Mean & Rubble & Aerial & Mean \\
\midrule
LangSplat~\cite{Qin_2024_CVPR}  & 0.3926 & 0.3011 & 0.3163 & 0.3421 & - & - & - \\
OpenGaussian~\cite{wu2024opengaussian} & 0.5755 & 0.4416 & 0.4357 & 0.5095 & - & - & - \\
Gaussian Grouping~\cite{gaussian_grouping} & 0.5551 & \textbf{0.5145} & 0.5945 & 0.5582 & - & - & - \\
\textbf{QueryGaussian (Ours)} & \textbf{0.7742} & 0.4764 & \textbf{0.5989} & \textbf{0.6380} & \textbf{0.7711} & \textbf{0.7625} & \textbf{0.7676} \\
\bottomrule
\end{tabular}}
\end{table*}

\begin{figure*}[!ht]
  \includegraphics[width=\textwidth]{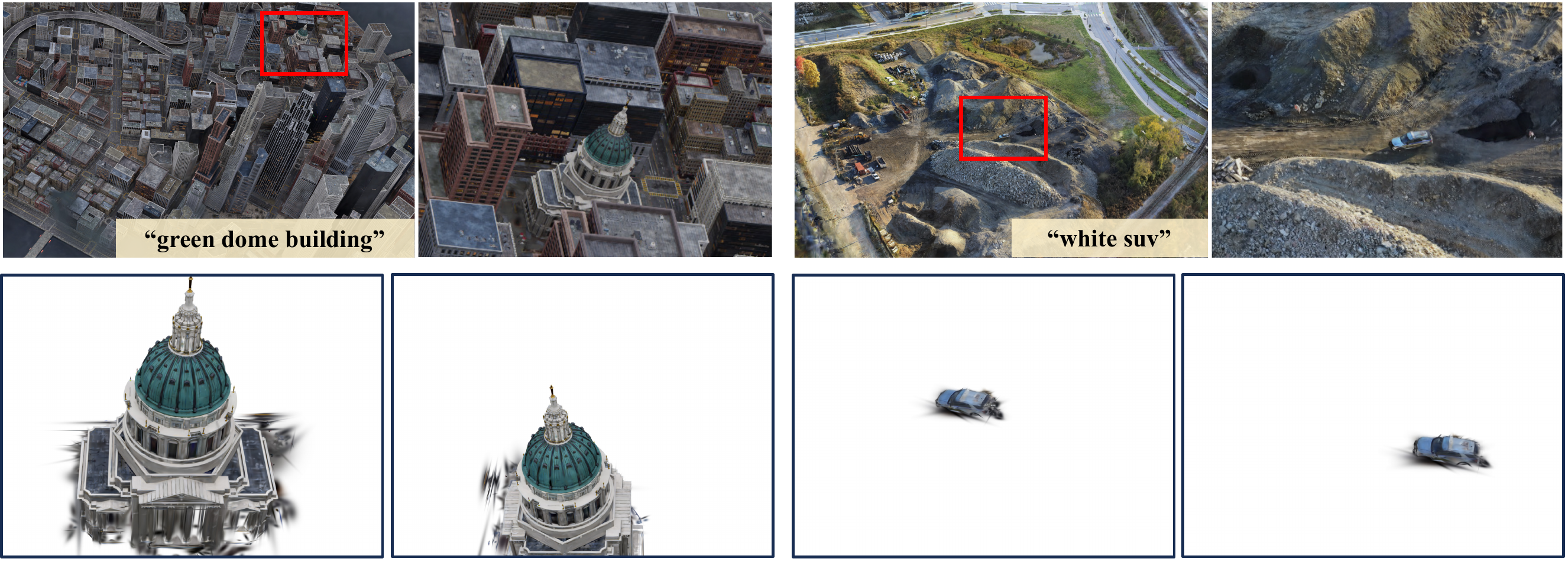}
  \caption{Qualitative results on large-scale outdoor scenes. Existing scene-level methods fail due to OOM, while QueryGaussian successfully localizes fine-grained targets such as buildings and vehicles.}
  \label{fig:lss}
\end{figure*}

\subsection{Instance Query on Arbitrary-Scale 3D Scenes}

\subsubsection{Small-Scale Indoor Scenes}

\begin{table*}[!ht]
\centering
\caption{Time and memory cost on small-scale indoor scenes.}
\label{tab:cost_results_lerf}
\resizebox{\textwidth}{!}{
\begin{tabular}{l|ccccc}
\toprule
Methods & Preprocess time & Training time & Query time & Total & GPU memory \\ 
\midrule
LangSplat~\cite{Qin_2024_CVPR}        & $>3$h      & $\sim$1h   & $\sim$60$s$ & $>4$h      & $\sim$15GB \\
OpenGaussian~\cite{wu2024opengaussian}& $>1$h      & $>1$h      & $\sim$60$s$ & $>1$h      & $\sim$24GB \\
Gaussian Grouping~\cite{gaussian_grouping}
                                      & $>12$h     & $\sim$1h   & $\sim$60$s$ & $>13$h     & $>26$GB \\
\textbf{QueryGaussian (Ours)}         & \textbf{0} & \textbf{0} & \textbf{$\sim$20$s$} & \textbf{20$s$} & \textbf{$\sim$7GB} \\
\bottomrule
\end{tabular}}
\end{table*}

On small-scale indoor scenes, QueryGaussian achieves the best mean IoU of 0.6380, outperforming Gaussian Grouping (0.5582) and OpenGaussian (0.5095), as shown in Table~\ref{tab:iou_results}. The gain is especially clear on Figurines and Teatime, where precise 2D-to-3D association produces cleaner boundaries and fewer floaters, consistent with the qualitative results in Figure~\ref{fig:lerf}.

The efficiency gap is also substantial. As shown in Table~\ref{tab:cost_results_lerf}, prior methods require expensive preprocessing or training before retrieval, whereas QueryGaussian is fully training-free and completes an end-to-end query in about 20 seconds using about 7GB VRAM. This corresponds to roughly $180\times$ faster end-to-end processing and more than 70\% lower memory usage than prior methods.

\begin{table*}[!htp]
\centering
\caption{Time and memory cost on large-scale outdoor scenes. All methods are tested on a single RTX 4090 (24GB). ``OOM'' denotes failure due to GPU memory limits.}
\label{tab:cost_results_lss}
\resizebox{\textwidth}{!}{
\begin{tabular}{l|ccccc}
\toprule
Methods & Preprocess time & Training time & Query time & Total & GPU memory \\
\midrule
LangSplat~\cite{Qin_2024_CVPR}              & \textit{OOM} & \textit{OOM} & - & - & $>>$24GB \\
OpenGaussian~\cite{wu2024opengaussian}      & \textit{OOM} & \textit{OOM} & - & - & $>>$24GB \\
Gaussian Grouping~\cite{gaussian_grouping}  & \textit{OOM} & \textit{OOM} & - & - & $>>$26GB \\
\textbf{QueryGaussian (Ours)}               & \textbf{0} & \textbf{0} & \textbf{$\sim$60s} & \textbf{$\sim$60s} & \textbf{$\sim$7GB} \\
\bottomrule
\end{tabular}
}
\end{table*}

\begin{figure*}[!htbp]
  \centering
  \includegraphics[width=\textwidth]{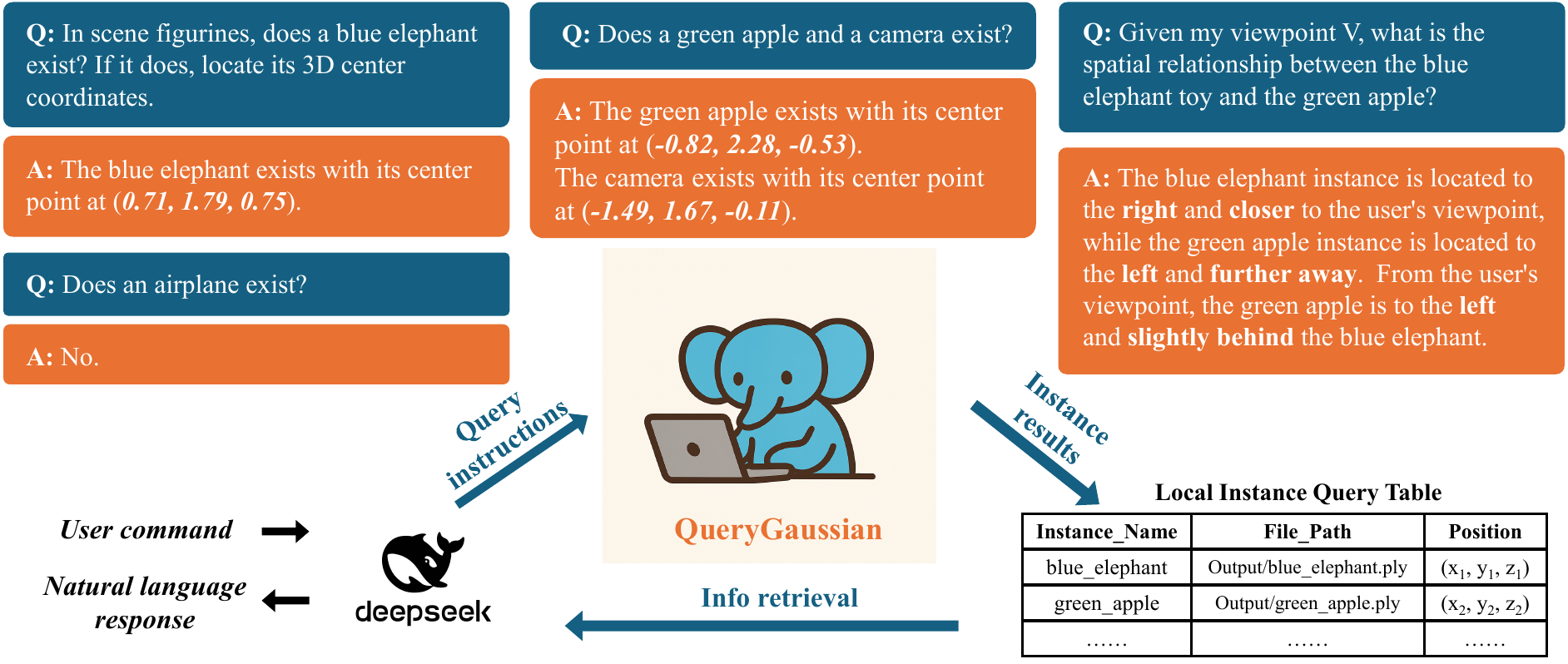}
  \caption{Overview of the 3D spatial reasoning agent. The LLM decomposes a user query into retrieval instructions, QueryGaussian returns masks and 3D coordinates, and the retrieved results are used for subsequent spatial reasoning.}
  \label{fig:3DQA}
\end{figure*}

\subsubsection{Large-Scale Outdoor Scenes}

We further evaluate QueryGaussian on large-scale outdoor scenes containing more than 10 million Gaussians. As shown in Table~\ref{tab:iou_results}, QueryGaussian reaches a mean IoU of 0.7676 on this setting. In contrast, all scene-level embedding baselines fail with OOM on a 24GB GPU, as reported in Table~\ref{tab:cost_results_lss}.

This failure is not incidental. Scene-level methods require storing or optimizing semantic features for the full scene, and their memory cost grows with the number of Gaussians. QueryGaussian avoids this bottleneck by performing retrieval only for the queried target in image space. As a result, it processes city-scale scenes in about 60 seconds with about 7GB memory. Figure~\ref{fig:lss} shows representative examples.

\subsubsection{Comparative Analysis}

The advantage of QueryGaussian comes from two design choices. First, the maximum-weight map is generated directly during rasterization, so 2D-to-3D indexing is obtained with negligible extra cost. Second, semantic processing is shifted to rendered images rather than full-scene 3D features. This changes the dominant semantic cost from scaling with the number of Gaussians to scaling with the number of image pixels.

These two properties explain the results in Tables~\ref{tab:cost_results_lerf} and \ref{tab:cost_results_lss}. Previous methods couple semantic understanding with scene-wide 3D optimization, which leads to high preprocessing cost and large memory usage. QueryGaussian instead performs direct query-time lifting and refinement, achieving competitive or better accuracy while remaining practical on consumer hardware.

\subsection{Multi-Instance Retrieval and Spatial Reasoning Agent}
\label{sec:spatial_reasoning}

To illustrate a downstream use case, we combine QueryGaussian with a reasoning-oriented large language model, DeepSeek-R1 \cite{deepseekai2025deepseekr1incentivizingreasoningcapability}, to build a simple 3D question-answering agent. As shown in Figure~\ref{fig:3DQA}, the LLM first decomposes a user query into executable instance retrieval requests. QueryGaussian then returns the corresponding masks and 3D centroids, which are stored in a local instance table. The LLM uses this retrieved geometric information to answer spatial questions.

Preliminary examples show that this modular pipeline can handle multi-instance spatial queries, such as relative position reasoning between objects. Since QueryGaussian is training-free and efficient, it supports this interaction without additional scene-specific optimization. At the same time, the current system is limited to visually grounded reasoning and does not model non-visual physical properties.

\section{Conclusion}

We propose \textbf{QueryGaussian}, a training-free framework that redefines scalable 3D instance retrieval via an on-demand strategy. By decoupling semantic understanding from scene representation, our method enables expeditious retrieval on city-scale scenes ($>10$ million Gaussians) using consumer hardware—overcoming the memory failures inherent in prior scene-level embedding paradigms. While pre-computed embeddings may benefit high-frequency queries in small static settings, QueryGaussian is uniquely advantageous for city-scale applications where pre-training is memory-prohibitive (OOM), cold-start scenarios requiring immediate interaction, and dynamic environments where cached features would effectively become invalid. This work advances practical multimedia retrieval, facilitating future research in native multi-instance querying and physics-aware reasoning.


\bibliographystyle{plain}
\bibliography{refs}

\end{document}